\documentclass[conference]{IEEEtran}
\IEEEoverridecommandlockouts
\usepackage{cite}
\usepackage{amsmath,amssymb,amsfonts}
\usepackage{algorithmic}
\usepackage{graphicx}
\usepackage{textcomp}
\usepackage{booktabs, times, epsfig, graphicx, amsmath, amssymb,url, multirow,comment,commath}

\graphicspath{ {models/} }

\begin{document}

\title{Collaborative Filtering with Label Consistent Restricted Boltzmann Machine}

\author{\IEEEauthorblockN{Sagar Verma}
IIIT Delhi\\
sagar15056@iiitd.ac.in
\and
\IEEEauthorblockN{Prince Patel}
IIIT Delhi\\
prince15046@iiitd.ac.in
\and
\IEEEauthorblockN{Angshul Majumdar}
IIIT Delhi\\
angshul@iiitd.ac.in
}

\IEEEpubid{\makebox[\columnwidth]{978-1-5386-2241-4/17/\$31.00~\copyright~2017 IEEE \hfill} \hspace{\columnsep}\makebox[\columnwidth]{ }}

\maketitle

\IEEEpubidadjcol

\begin{abstract}
The possibility of employing restricted Boltzmann machine (RBM) for collaborative filtering has been known for about a decade. However, there has been hardly any work on this topic since 2007. This work revisits the application of RBM in recommender systems. RBM based collaborative filtering only used the rating information; this is an unsupervised architecture. This work adds supervision by exploiting user demographic information and item metadata. A network is learned from the representation layer to the labels (metadata). The proposed label consistent RBM formulation improves significantly on the existing RBM based approach and yield results at par with the state-of-the-art latent factor based models.
\end{abstract}


\section{Introduction}
With the worldwide boom of E-Commerce (business-to-client) research in recommender systems has become one of the top priorities both for academia and the industry \cite{Schafer1, Schafer2}. Recommender systems are beneficial for both, the business, and the client. Unlike a physical marketplace, online portals have virtually an inexhaustible collection of items. It is a huge task for the user to sift through all the options and buy/rent the one he/she is satisfied with. Thus, the recommender systems assist the buyer/client with tailored suggestions. For the E-Commerce portal, without proper recommendations the user does not purchase items; hence the portal looses on business opportunity and thereby loses prospective revenue.
The initial days of recommender systems saw the application of content based filtering to recommender systems \cite{Meteren1}. Content based filtering was a well-established approach in information retrieval – this was in the early 2000s. However, it never became popular; several reasons are briefly mentioned in \cite{Hofmann1}. Mainly owing to the requirement of user intervention in the definition of ‘content’; i.e. one needed to find out the exact attributes for matches between user and item. For example, for books the factors might be the author, genre, publisher; for music, they might be the singer, genre; for movie, they can be anything ranging from the actors to the director to the production house to the genre.
One did not know if the expert defined list had already captured all the possible variabilities in the chosen attributes. If the list of attributes is too small, important factors would be missing; if the list is too large one might end up capturing noise in the data.
During the same time (the early 2000s) more abstract yet powerful techniques based on latent factor modeling and representation learning started gaining momentum. Instead of explicitly designing the attributes, latent factor model makes an abstract assumption. It assumes that the user’s choice of items is guided by several latent factors. Instead of designing these factors (as in content based filtering), they were learned from the data. Eventually, matrix factorization \cite{Koren, Rennie} became the most popular technique for latent factor model especially after the announcement of the famed Netflix competition \cite{Bennett}.
In the late 90’s and early 2000’s neighborhood based approaches gained popularity in collaborative filtering. They were interpolation based techniques. In the user-based approach \cite{Herlocker}, similar users were selected to constitute the ‘neighbourhood’, and the ratings of these users were used to impute the missing values. The same could be done from an item perspective \cite{Sarwar}. These approaches were simple and easy to interpret. But they were heuristic, the interpolation weights were defined rather arbitrarily. The neighborhood based method yielded significantly lower accuracy (at least on the benchmark databases) than the more powerful albeit abstract latent factor models.
Even though the matrix factorization technique was the popular choice for latent factor based collaborative filtering, a seminal work \cite{Salakhutdinov} showed the possibility of using another representation learning/latent factor approach for collaborative filtering; it was the restricted Boltzmann machine (RBM). There is hardly any work on RBM based collaborative filtering since the publication of \cite{Salakhutdinov}.
The basic RBM formulation used in \cite{Salakhutdinov} is an unsupervised one; it was only based on the user’s rating on the items. However, in a real system, users’ and items’ metadata is always available. User demographic information such as age, occupation, gender etc. is collected as a part of the sign-up process. The item information is also collected during its registration. Prior RBM based formulation could not make use of such auxiliary formation. In this work, we propose to improve upon \cite{Salakhutdinov} (in terms of rating prediction accuracy) by exploiting the user demographics and item metadata.
In recent times (last few years), researchers have started exploring the possibility of using another powerful representation learning technique for collaborative filtering, stacked autoencoder. Both stacked autoencoders and deep belief network (built from layers of RBM) are used to train deep neural networks. Almost all studies in autoencoder based collaborative filtering are minor variations of each other. The basic autoencoder formulation is directly used in \cite{Strub1,Ouyang2014,Wu}. In \cite{Strub2} baseline prediction is used along with the ratings in the autoencoder framework; the baseline values are simply appended with the available ratings so that the autoencoder learns to reconstruct both the ratings and the baseline values. A combination of marginalized denoising autoencoder and probabilistic matrix factorization is used in \cite{Li} for rating prediction.
All representation learning approaches are inherently nonconvex. Therefore, the associated theoretical problems are ever present. There is an elegant convex solution to the matrix factorization approach for collaborative filtering; this is called matrix completion \cite{Abernethy,pmlr-v19-shamir11a}. It is a convex variant which directly solves for the missing ratings instead of going through the intermediate steps of determining the latent factors for the users and the items. However, this is unrelated to the representation learning/latent factor model based approaches we will not discuss it in detail.
The rest of the paper will be organized into several sections. Background on collaborative filtering will be discussed in the next section. The proposed formulation will be detailed in section 3. The experimental results will be shown in section 4. Conclusions of this work and future directions will be discussed in section 5.

\section{Background}

\subsection{Neighborhood Models}
Even though neighborhood/memory based models are not the focus of this work, we discuss it nevertheless for the sake of completion.
Collaborative filtering can be thought of as a matrix completion problem. We assume that the users are the rows and the items along the columns. Each user has rated a few items. Based on such parsimonious ratings, recommender systems need to predict the missing ratings. Once it has predicted the ratings, it recommends items to users with high valued ratings.
Neighborhood based approach follows a simple interpolation based formulation. For an active user, it predicts the missing ratings. First, it finds users similar to the active user (neighborhood) by computing some kind of similarity (cosine, inverse distance etc.). Next, it interpolates the active user’s missing ratings as a linear combination of the ratings from the neighborhood. The linear interpolation weights are heuristically fixed; usually, they are the normalized similarity weights computed in the first step.
Such a technique is called the user-based approach \cite{Herlocker}. One can perform exactly the same steps from an items perspective, leading to the item based approach \cite{Sarwar}. There are combined user and item based techniques \cite{Wang} as well. Such neighborhood based models are simple to understand and implement. The results are easy to analyze. However, they are heuristic and do not yield very good results.

\subsection{Matrix Factorization}
The latent factor model assumes that the user’s choice in items is determined by a handful of factors. For example, in content based filtering, these factors are hand-picked; for books, they might be author, publisher, and genre and for TV shows they might be star cast and genre. However latent factor models do not try to select these factors, rather they learn the abstract hidden factors from the data. Say a user’s latent factor be represented as a vector $u_i$ and the item’s latent factor be represented as $v_j$. The user will ‘like’ the item and provide a high rating if the corresponding latent factors match. This is captured as an inner product between the two vectors. Therefore, the rating of the $i^{th}$ user on the $j^{th}$ item is modeled
as:

\begin{equation}
  r_{i,j} = u_iv_j, \forall i,j
\end{equation}

For the entire rating matrix, i.e. for $M$ users and $N$ items, the latent factor model is expressed as:

\begin{equation}
  R = UV \text{ where } U = [u+1 \abs{\dots} u_M] \text{ } V^T = [v_1 \abs{\dots} v_N]
\end{equation}

In collaborative filtering, the full rating matrix is not available. A partially sampled version (the items for which users have provided ratings) are only available. The objective is to predict the missing values so that recommendations can be made based on those. The rating acquisition can be
mathematically expressed as:

\begin{equation}
  Y = M \odot R = M \odot (UV)
\end{equation}

Here $M$ is the binary mask; it has one where the ratings are available and zeroes where they are missing. The symbol $\odot$ shows binary multiplication.

This (3) is a highly-under-determined problem. Usually, less than 5\% of the matrix is filled. Therefore, correctly predicting the missing ratings is a challenging problem. However, the number of latent factors are usually small, for books it may be just two or three, for movies it can be as large as 40; but still, the number of factors are much smaller than the rows or columns of the matrix. Hence the rating matrix will be a low-rank matrix. Therefore, the number of parameters to be learned (elements in $U$ and $V$) are usually much smaller than the dimensionality of the matrix. Intuitively speaking, this gives some hope for solving the under-determined system (3); as long as the pieces of information (ratings in this case) is larger than the number of parameters to be learned, we can hope to be able to estimate them.
The user and item latent factor matrices are recovered by solving the standard matrix factorization problem

\begin{equation}
  \min_{U,V} \norm{Y-M \odot (UV)}^2_F + \lambda (\norm{U}^2_F+\norm{V}^2_F)
\end{equation}

The Tikhonov type penalties prevent overfitting. There can be various ways to solve the factorization problem (4). It can be as simple as stochastic gradient descent and alternating least squares to as complex as probabilistic matrix factorization.
Authors in \cite{Gogna1} argued that even though user’s latent factors can be dense since it expected that human beings have a certain degree of affinity towards all factors; an item cannot possess all the latent factors simultaneously. Hence, it is likely that they would be sparse. Following this argument, a sparsity promoting l1-norm penalty on V has been proposed in the aforesaid study.

\begin{equation}
  \min_{U,V} \norm{Y-M \odot (UV)}^2_F + \lambda (\norm{U}^2_F+\norm{V}_1)
\end{equation}

The problems (4 and 5) are bi-linear, hence non-convex. There is no guarantee that they will converge to the optimum. Matrix completion is an elegant formulation that directly solves for the ratings and not the user and the item latent factors. It recovers the rating matrix directly by searching for
a low-rank solution. However, minimizing the rank (number of singular values) is a NP-hard problem hence it's closest convex proxy the nuclear norm (sum of singular values) is used instead \cite{Candes,Recht}. The formulation for matrix completion is,

\begin{equation}
  \min_{R} \norm{Y-M \odot R}^2_F + \lambda (\norm{R}_NN)
\end{equation}

This is a convex formulation which can be solved by semidefinite programming. Many faster algorithms also exist. 
As mentioned earlier, collaborative filtering is a highly under determined problem. So far, we only described basic techniques which required ratings only. In such a scenario, it is likely that using secondary information may improve the results. In practice metadata regarding the user is easily available; demographic information of the users is also available to the portal from the registration / sign-up process. Several studies \cite{Gogna3,Gu} have used this auxiliary information to boost the accuracy of a recommender system. Others have combined information from neighborhood based models with matrix factorization. In \cite{pmlr-v9-sutskever10a} the similarity between individuals was used as a graph regularization in the matrix factorization framework.

  \subsection{Restricted Boltzmann Machine}

  \begin{figure}
    \centering
      \includegraphics[width=0.2\linewidth]{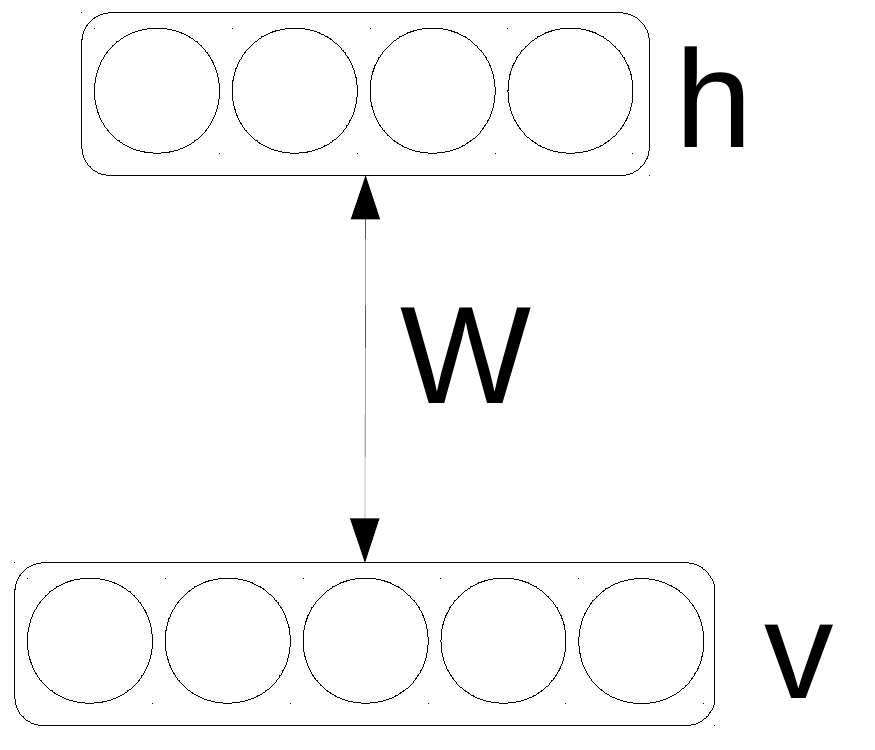}
        \caption{Restricted Boltzman machine}
        \label{fig:3}
  \end{figure}

The restricted Boltzmann machine has been proposed in \cite{Salakhutdinov} to address the collaborative filtering problem. It is a two-layered (input and representation) undirected graphical model (as shown in Figure 1).

A simple RBM has a visible layer $V$ and hidden layer $h$. Edges connecting these two layers are undirected and the whole network is a bipartite graph. The visible layer $V$ is $K \times m$ matrix with $V_i^k=1$, if a user rating for movie $i$ is $k$. Hidden layer $h$ is a $1 \times F$ vector. Joint distribution between hidden layer $h$ and visible layer $V$ takes the form:

  \begin{equation} 
    p(V,h) \propto e^{-E(V,h)}
    \label{eq:4}
  \end{equation}

where $E(V,h) = -h^T W V - a^T V - b^T h W$ with parameters $\theta = (W,a,b,)$. Equation \ref{eq:5} captures predictive information about the input vector. $P(h|V)$ has a similar form.

  \begin{equation}
    p(V_i^k = 1|h) = \frac{exp(V_i^k + \sum_{j=1}^{F}h_j W_{ij}^k)}{\sum_{l=1}^{K}exp(V_i^l+\sum_{j=1}^{F}h_jW_{ij}^l)}
    \label{eq:5}
  \end{equation}

  \begin{equation}
    p(h_j=1|V) = \sigma(b_j+\sum_{i=1}^{m}\sum_{k=1}^{K}V_i^kW_{ij}^k)
    \label{eq:6}
  \end{equation}

where $\sigma(x)$ is a logistic function, $W_{ij}^K$ is weight parameter between hidden unit(feature) $j$ and rating $k$ of movie $i$, $b_i^k$ is the bias of rating $k$ for movie $i$, $b_j$ is the bias of feature $j$.

The marginal distribution over the visible ratings $V$ is:

  \begin{equation}
    p(V) = \sum\frac{exp(-E(V,h))}{\sum_{V'h'}exp(-E(V',h'))}
    \label{eq:7}
  \end{equation}

  Here $E(V,h)$ is energy function given by Equation \ref{eq:4}.

  To update parameters gradient ascent is required in the log-likelihood, this can be obtained from equation \ref{eq:7}.

  \begin{equation}
    \Delta W_{ij}^k = \mu \frac{\delta logp(V)}{\delta W_{ij}^k} = \mu (<V_i^kh_j>_{data} - <V_i^kh_j>_{model})
    \label{eq:8}
  \end{equation}

  where $\mu$ is the learning rate. The expectation $<V_i^kh_j>_{data}$ defines the frequency with which movie rating $k$ and feature $j$ are on together when the features are being driven by the observed user-rating data from the training set using Equation \ref{eq:7}, and $<.>_{model}$ is an expectation with respect to the distribution defined by model. The expectation $<.>_{model}$ cannot be computed analytically in less than exponential time. Contrastive Divergence (CD) \cite{pmlr-v9-sutskever10a} is used to approximate computation of gradient of objective function.

  \begin{equation}
    \Delta W_{ij}^k = \mu (<V_i^kh_k>_{data} - <V_i^kh_k>_T)
    \label{eq:9}
  \end{equation}

  The expectation $<.>_{data}$ represents a distribution of samples from running the Gibbs sampler, initialized at the data, for $T$ full steps.

  Given the observed ratings $R$, we can predict a rating for a new query movie $m$ in time linear in the number of hidden units:

  \begin{align}
    & p(V_m^k=1|R)  \propto \sum_{h_1\dots h_p} exp(-E(V_m^k,R,h)) \nonumber \\
    &              \propto \Gamma_m^k \Pi_{j=1}^F \sum_{h_j \in \{0,1\}} exp(\sum_{il}V_i^lh_jW_{ij}^l + V_m^kh_jW_{mj}^k + b_jh_j) \nonumber \\
    &              = \Gamma_m^k \Gamma_{j=1}^F (1+exp(\sum_{il}V_i^lW_{ij}^l+V_m^kW_{mj}^k+b_j)) 
                 \label{eq:10}
  \end{align}

  where $\Gamma_m^k = exp(V_m^kb_m^k)$. 

  Once the unnormalized scores are obtained, pick the rating with the maximum score as the prediction, or perform normalization over $K$ values to get probabilities $p(V_m=k|R)$ and take the expectation $E[V_m]$ as the prediction.

\section{Proposed Approach} 

  We will first discuss prior studies in label consistency penalties in latent factor model. This was introduced by one of the authors in prior studies \cite{gogna2016supervised}. In the next sub-section, we will describe our proposed formulation.

  \subsection{Label Consistent Latent Factor Model}
   The basic formulation for the latent factor model is as follows, 

  \begin{equation}
    Y = M \odot R = M \odot (UV)
  \end{equation}

The symbols have their usual meaning. In the label consistent formulation \cite{gogna2016supervised}. The user latent factors are mapped to demographic information encoded as binarized labels and the item metadata is linearly mapped to their respective binarized metadata information. Formation of the label information from metadata is explained with examples.

Let us take the example of movie recommendation. There are many features for the movies available movie id, movie title, release date, video release date, IMDb URL, Genre (Action, Adventure, Animation, Children’s, Comedy, Crime, Documentary, Drama, Fantasy, Film-Noir, Horror, Musical, Mystery, Romance, Sci- Fi, Thriller, War, Western). All except the genre information is irrelevant for choosing a movie. We have considered only the genre information here.

\begin{figure}
    \centering
      \includegraphics[width=0.6\linewidth]{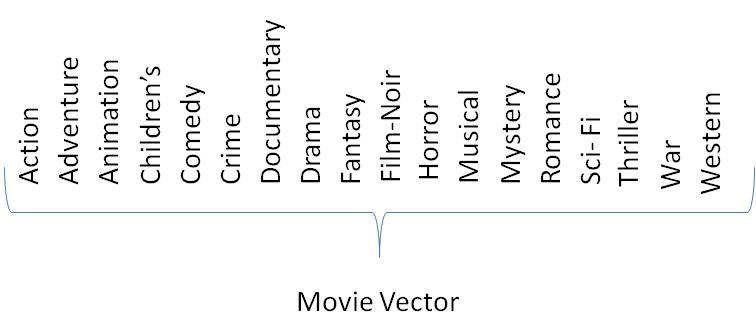}
        \caption{Movie(genre) feature vector}
        \label{fig:genre_vector}
  \end{figure}

First, we will describe how the movie features have been-generated. We generate a binary vector from the genre, the vector contains a 1 if the movie belongs to that genre or 0 otherwise. The vector is shown in Figure \ref{fig:genre_vector}.

Say a movie like Shawshank Redemption is tagged as ‘crime’ and ‘drama’ in IMDB. The corresponding feature vector will be [0,0,0,0,0,1,0,1,0,0,0,0,0,0,0,0,0,0]T; It has 1’s corresponding to crime and drama and 0 everywhere else. Corresponding to each movie/item one has such a label vector. For all the movies let it be denoted as Q. 

We now discuss encoding the user demographic information. Encoding the gender information is the simplest. It is a tuple – encoded as [1,0]T for male and [0,1] T for female.

\begin{figure}
    \centering
      \includegraphics[width=0.6\linewidth]{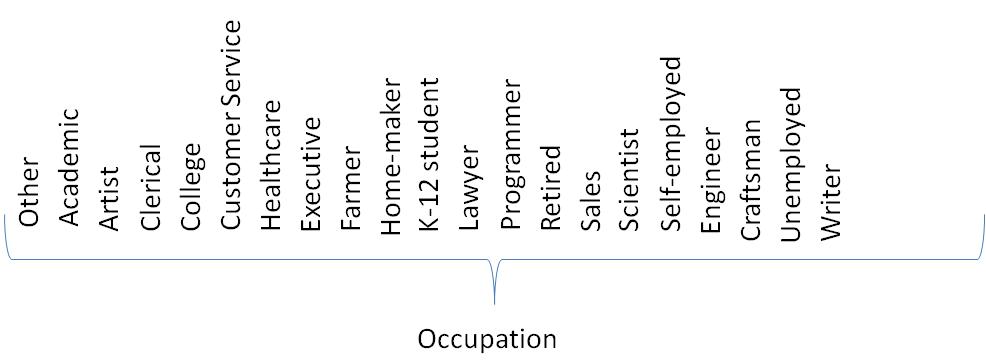}
        \caption{Encoding Occupation Information}
        \label{fig:occ_vector}
  \end{figure}

To encode the occupation information, we have an ordered representation of the different occupations as in Figure \ref{fig:occ_vector}. For a particular user, one of the occupations is 1 the rest are 0s.
  
To encode the age information, we divide the users into several ranges; more specifically into set into 8 groups (7-14,14-21, 22-28, 29-36, 37-48, 49-55, 56-65 and 66-73). The groups are divided keeping in mind the relevance of mentality of users according to age. The particular age group, where the individual belongs to is 1 and the rest of them are 0s.

Thus, for ever user one will have a binary label vector. For all users, this is a matrix T.

When incorporated in the latent factor model \cite{gogna2016supervised} using the matrix factorization framework, the learning is expressed as,

\begin{equation}
  \min_{U,V,M_1,M_2} \norm{Y-M \odot (UV)}^2_F | \delta(\norm{T - M_1 U}^2_F + \norm{Q - M_2 V}^2_F)
\end{equation}

Instead of learning the maps from the latent factors, one can also learn it directly from the ratings \cite{gogna2016supervised}, i.e. the user’s ratings are linearly mapped to the user’s binary labels encoding the demographic information and the item ratings are mapped to their corresponding binarized metadata. This is expressed as,

\begin{equation}
  \min_{U,V,M_1,M_2} \norm{Y-M \odot (R)}^2_F | \delta(\norm{T - M_1 R}^2_F + \norm{Q - R M_2}^2_F)
\end{equation}

Collaborative filtering is a highly-under-determined problem. We have mentioned before, that only 5\% of the ratings are available and one needs to predict the remaining 95\%. In such a situation, any extra information helps. It is not surprising that the metadata information in \cite{gogna2016supervised} indeed improve the results. 

    \subsection{Label Consistent RBM}

  Motivated by the success of label consistent formulations in latent factor models, we propose a novel label consistent RBM to improve upon the existing CF formulation \cite{Salakhutdinov}. However, unlike \cite{gogna2016supervised}, it is not possible to capture both user and item metadata in a single framework for the RBM based formulations. To capture item metadata we add genre information of movies to the simple RBM. In this variation, there are two different layers of visible units, one layer consists of ratings of N users and another layer is of binary label information (genre vector Q) for that movie.

    \begin{figure}
      \centering
      \includegraphics[width=0.3\linewidth]{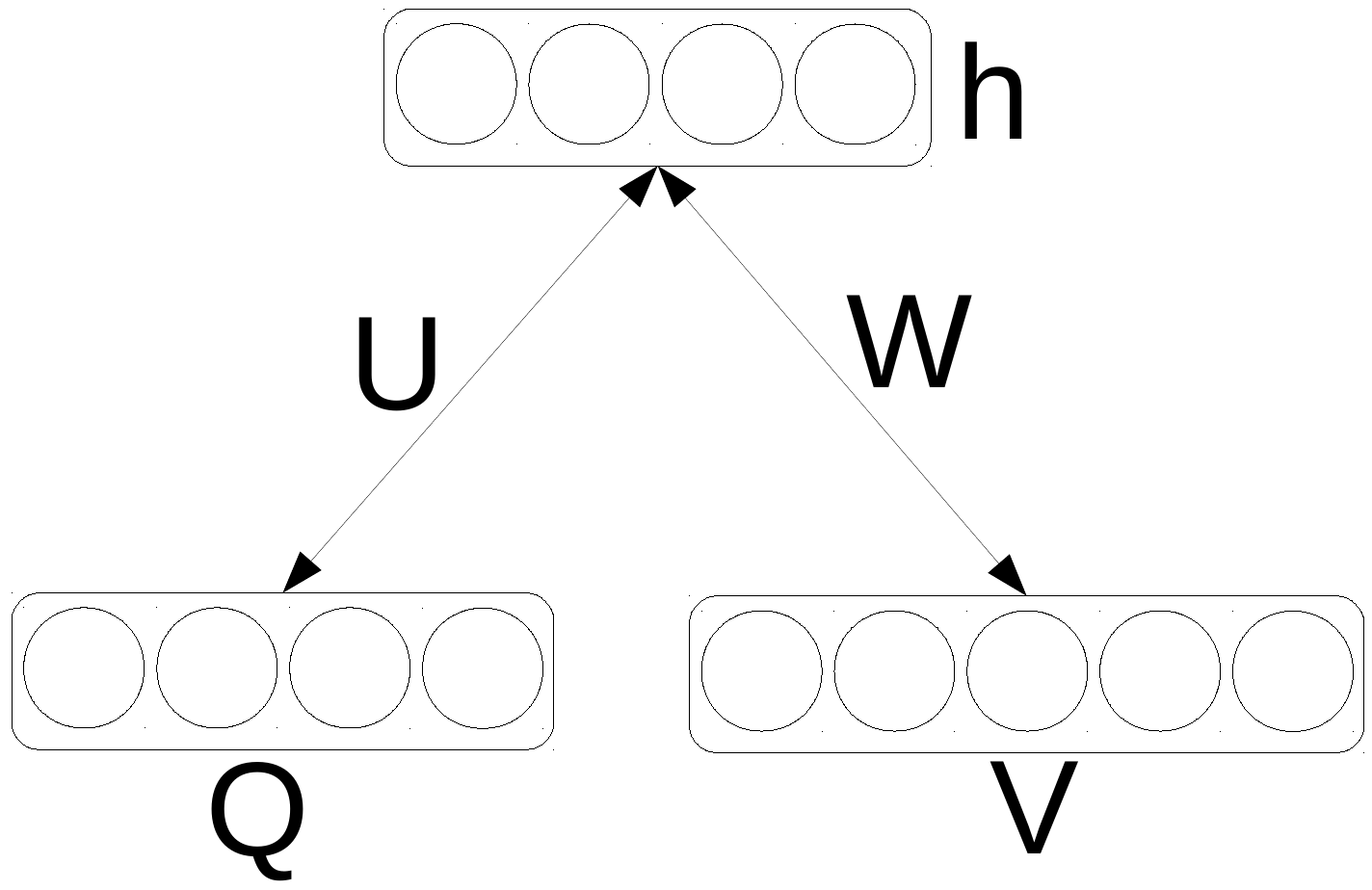}
      \caption{RBM variant with item metadata}
      \label{fig:4}
    \end{figure}

    The joint probability distribution of this model is given by,

    \begin{equation}
      p(Q,V,h) \propto e^{-E(Q,V,h)}
      \label{eq:11}
    \end{equation}

    where we define the new energy function as follows:

    \begin{equation}
      E(Q,V,h) = -h^TWV - a^TV -b^Th - c^TQ - h^TUQ
      \label{eq:12}
    \end{equation}

    with parameters $\Theta = (W,a,b,c,U)$. The model is illustrated in Figure \ref{fig:4}. We find the values of visible and hidden units using Equations \ref{eq:6}, \ref{eq:13} and \ref{eq:14} respectively.

    \begin{equation}
      p(h_j=1|V,Q) = \sigma(b_j+U_{jq}+\sum_qW_{jq}V_q)
      \label{eq:13}
    \end{equation}

    \begin{equation}
      p(Q_q=1|h) = \frac{exp(c_q+\sum_{j}U_{jq}h_j)}{\sum_{q=1}^{Q}exp(c_q+\sum_{j}U_{jq}h_j)}
      \label{eq:14}
    \end{equation}

    where $\sigma$ is the logistic sigmoid. These equations are meant to capture the predictive information about the input vector as well as the target class.

    To perform the learning, we use the same Equation \ref{eq:9}, representation is as follows:

    \begin{align}
      \Delta W_{ij}^k & = \mu \frac{\delta logp(V,Q)}{\delta W_{ij}^k} \nonumber \\ 
                      & = \mu (<V_i^kh_jQ_i>_{data}-<V_i^kh_jQ_i>_{model})
                      \label{eq:15}
    \end{align}

    where $\mu$ is learning rate.

    \begin{figure}
      \centering
      \includegraphics[width=0.5\linewidth]{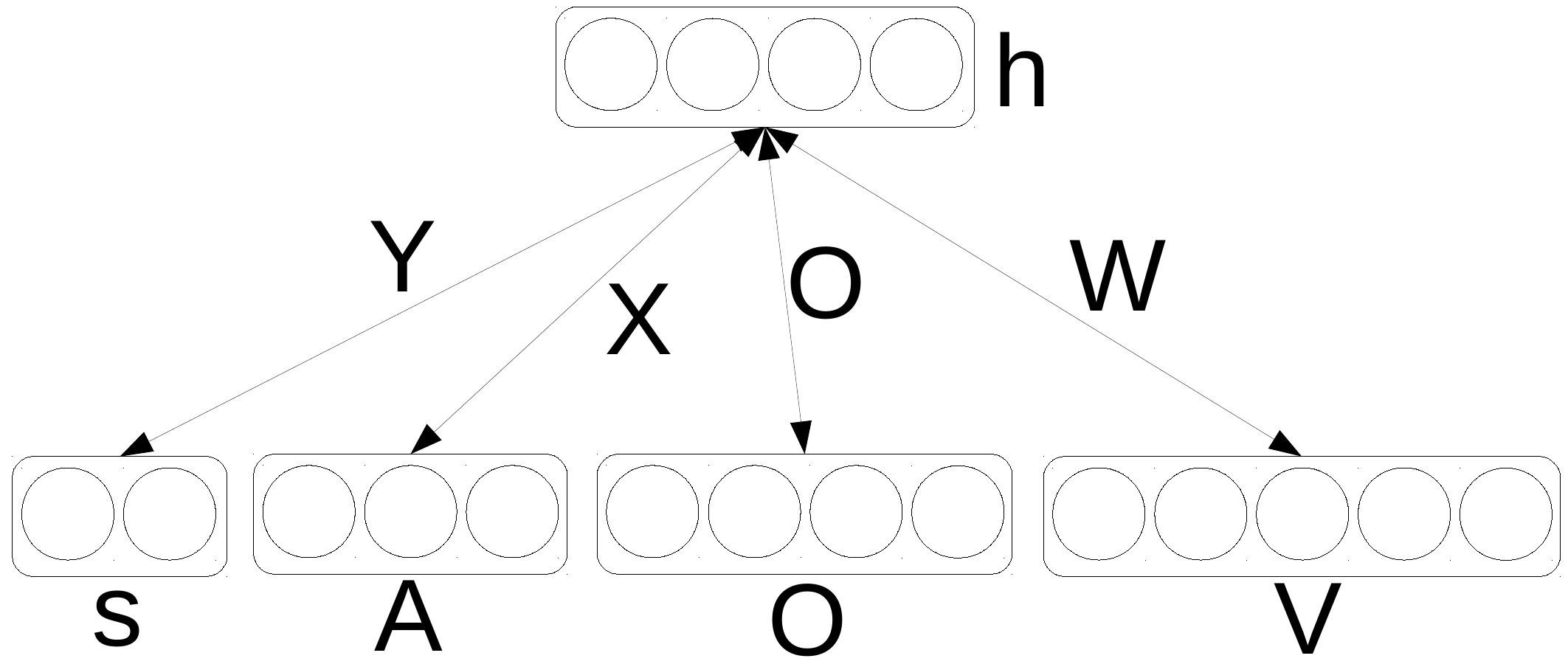}
      \caption{RBM variant with user metadata}
      \label{fig:5}
    \end{figure}

    To capture user meta data we add occupation, age, and gender to the simple RBM. In this variation, we have four different visible layers. Layer 1 consists of ratings given to all the movies by user $i$, and has size $M \times K$. Layer 2 captures occupation information for user $i$, and has size $O$. Layer 3 captures age information for user $i$, and has size $A$. Layer 4 captures gender information for user $i$, and has size $S$. The joint probability distribution of this model is given by:

    \begin{equation}
      p(O,S,A,V,h) \propto e^{-E(O,S,A,V,h)}
      \label{eq:16}
    \end{equation}

    The modified energy function now becomes:

    \begin{align}
      E(O,S,A,V,h) &= -h^tWV-a^TV-b^Th-d^TO \nonumber \\
      &-d^TZ-e^TA-e^TXA-f^TS-f^TYS
      \label{eq:17}
    \end{align}

    With parameters $\Theta = (W,a,b,d,e,f,X,Z,Y)$. This model is illustrated in Figure \ref{fig:5}. To compute the parameters, we use the following equations:

    \begin{equation}
      p(h_j=1|V,O,S,A) = \sigma(b_j+Z_{jo}+X_{ja}+Y_{js}+\sum_{i}W_{ji}V_{i})
      \label{eq:18}
    \end{equation}

    For occupation of the user

    \begin{equation}
      p(O_o=1|h) = \frac{exp(d_o+\sum_jZ_{jo}h_j)}{\sum_{o=1}^Oexp(d_o+\sum_{j}Z_{jo}h_j)}
      \label{eq:19}
    \end{equation}

    For gender of the user, following equation is used

    \begin{equation}
      p(A_a=1|h) = \frac{exp(e_a+\sum_jX_{ja}h_j)}{\sum_{a=1}^Aexp(f_s+\sum_jY_{js}h_j)}
      \label{eq:20}
    \end{equation}

    For age of the user, following equation is used
    \begin{equation}
      p(S_s=1|h)=\frac{exp(f_s+\sum_jY_{js}h_j)}{\sum_{s=1}^Sexp(f_s+\sum_jY_{js}h_j)}
      \label{eq:21}
    \end{equation}

    To perform the learning, we use the same Equation \ref{eq:9}, representation is as follows:

    \begin{align}
      \Delta W_{ij}^k & = \mu \frac{\delta logp(V,O,S,A)}{\delta W_{ij}^k} \nonumber \\
                      & = \mu (<V_i^kO_iS_iA_ih_j>_{data} - <V_i^kO_iS_iA_ih_j>_{model})
                      \label{eq:22}
    \end{align}
    where $\mu$ is learning rate.

\section{Expertimental Evaluation} 

  \subsection{Description of dataset}

    Experiments were carried out on the popular Movielens dataset \cite{Harper}. We used the 100K and 1M dataset. These are the only datasets consisting of the user and item metadata. The larger 10M dataset does not contain this (user and item metadata) information.

    The 100K consists of 100,000 ratings (1-5) from 943 users and 1682 movies. Simple demographic information of the user including age, occupation, gender, and zip code is available. The zip code information is irrevelant\cite{Gogna3} and is not used here. For items genre information is present; other information can be obtained from IMDB, however, we do not use it in our experiments. In the 1M dataset, there are 1,000,209 ratings of approximately 3900 movies by 6040 users.

  \subsection{Experimental setup and evaluation criteria}

The standard experimental protocol is followed. 5 fold cross validation is performed on the standard splits.

In \cite{Gogna1} it was argued that the factors should be sparse. This is based on the observation that it is not possible for one item to possess all types of factors. Similarly, we argue that as users, we have an affinity towards a few factors only. Therefore, user latent factors should be sparse versions of our LC-RBM formulation. 

The learning rate is 0.0005 and number of epochs were 100. The number of hidden units is 100.

The quality of prediction is compared in terms of the standard metrics of Mean Absolute Error (MAE) and Root Mean Squared Error (RMSE)

\subsection{Results}

We have compared our technique with the baseline RBM \cite{Salakhutdinov}. For benchmarking, the de facto standard of probabilistic matrix factorization (PMF) \cite{Salakhutdinov2,Shan2,Ma} is used. The results are shown in the following table.

\renewcommand{\tabcolsep}{0.2cm}
    \begin{table}[!t]
        \centering
        \begin{tabular}{l  c  c  c  c}
            \toprule[0.2mm]
            \textbf{Method} & \multicolumn{2}{c}{\textbf{100K Dataset}} & \multicolumn{2}{c}{\textbf{1M Dataset}} \\
             & MAE & RMSE & MAE & RMSE \\
            \midrule
            LC-RBM (Item) & 0.7511 & 0.9451 & 0.7056 & 0.8625 \\
            Sparse LC-RBM (Item) & 0.7352 & 0.9207 & 0.6895 & 0.8521 \\
            LC-RBM (User) & 0.7709 & 0.9686 & 0.7269 & 0.8853 \\
            Sparse LC-RBM (User) & 0.7418 & 0.9317 & 0.6995 & 0.8721 \\
            RBM [12] & 0.8264 & 1.453 & 0.7821 & 1.023 \\
            PMF & 0.7564 & 0.9639 & 0.7241 & 0.9127 \\
            LCMC & 0.7193 & 0.9145 & 0.6731 & 0.8559 \\
            \bottomrule[0.2mm]
        \end{tabular}
        \caption{Comparitive Results}
        \label{table:1}
    \end{table}

Results show that our proposed formulation of supervised RBM indeed improves upon the unsupervised one \cite{Salakhutdinov}. Even without sparsity, we (LC-RBM item) produce better results than the benchmark RMF; the formulation using user metadata (without sparsity) is however slightly worse than PMF. With sparsity, our results improve even further. We always beat PMF. We do not beat the results from LCMC. To the best of our knowledge, it is the best-known algorithm for collaborative filtering today.
   
\section{Conclusion} 

Today RBM is popular as a building block for deep belief network (DBN). However, an early work on the topic showed, how it can be used in collaborative filtering. However the results from RBM could not compete with matrix factorization based latent factor models, and hence there is no significant work on this topic (RBM on recommender systems) since the publication of \cite{Salakhutdinov} in 2007. However, RBM based preprocessing has been done for improving results of matrix factorization; in fact, it has been used by the winners of the famous Netflix competition.

This work revisits RBM for collaborative filtering. We propose a new supervised model for RBM - label consistent RBM. This has been proposed to incorporate user and item metadata information commonly found in recommender systems. This significantly improves the performance, improving upon the state-of-the-art unsupervised techniques like probabilistic matrix factorization. In future, we would like to add graph-based similarity measure \cite{Lien} and item popularity information \cite{Steck} to our proposed method.

\bibliographystyle{IEEEtran}
\bibliography{ms} 

\end{document}